\documentclass{article}

\usepackage{microtype}
\usepackage{graphicx}
\usepackage{subfigure}
\usepackage{booktabs} 
\usepackage{mathtools}
\usepackage{eqnarray,amsmath,amssymb}
\usepackage{multicol}
\usepackage{blindtext}
\usepackage{caption}
\usepackage{filecontents}
\usepackage[cjk]{kotex}

\RequirePackageWithOptions{multicol}
\DeclareGraphicsExtensions{.pdf,.png,.jpg}

\usepackage{hyperref}

\usepackage[accepted]{icml2019}

\icmltitlerunning{Amplifying Imitation Effect}

\begin{document}
	\begin{CJK}{UTF8}{mj}
\twocolumn[
\icmltitle{Amplifying the Imitation Effect for Reinforcement Learning of \\UCAV’s Mission Execution}



\icmlsetsymbol{equal}{*}

\begin{icmlauthorlist}

\icmlauthor{Gyeong Taek Lee}{to}
\icmlauthor{Chang Ouk Kim}{to}
\end{icmlauthorlist}

\icmlaffiliation{to}{Department of Industrial Engineering, University of Yonsei, Seoul, Korea}

\icmlcorrespondingauthor{Chang Ouk Kim}{kimco@yonsei.ac.kr}


\icmlkeywords{Reinforcement Learning,Self-Imitation Learning,Random Network Distilliation}

\vskip 0.3in
]



\printAffiliationsAndNotice{\icmlEqualContribution} 

\begin{abstract}
This paper proposes a new reinforcement learning (RL) algorithm that enhances exploration by \textit{amplifying the imitation effect} (AIE). This algorithm consists of self-imitation learning and random network distillation algorithms. We argue that these two algorithms complement each other and that combining these two algorithms can amplify the imitation effect for exploration. In addition, by adding an intrinsic penalty reward to the state that the RL agent frequently visits and using replay memory for learning the feature state when using an exploration bonus, the proposed approach leads to deep exploration and deviates from the current converged policy. We verified the exploration performance of the algorithm through experiments in a two-dimensional grid environment. In addition, we applied the algorithm to a simulated environment of unmanned combat aerial vehicle (UCAV) mission execution, and the empirical results show that AIE is very effective for finding the UCAV's shortest flight path to avoid an enemy’s missiles.

\end{abstract}

\section{Introduction}
\label{submission}
Reinforcement learning (RL) aims to learn an optimal policy of the agent for a control problem by maximizing the expected return. RL shows high performance in dense reward environments such as games  \cite{mnih2013playing}. However, in many real-world problems,  rewards are extremely sparse, and in this case, it is necessary to explore the environment. The RL literature suggests exploration methods to solve this challenge, such as count-based exploration \cite{bellemare2016unifying,ostrovski2017count}, entropy-based exploration \cite{haarnoja2017reinforcement,ziebart2010modeling} and curiosity-based exploration  \cite{silvia2012curiosity,pathak2017curiosity,burda2018large,haber2018learning}.
 In recent years, many researchers have added an exploration bonus, often called curiosity or intrinsic reward, which is the difference between the predicted state and actual next state. The intrinsic reward is very efficient in exploration because the network for predicting the next state drives the agent to behave unexpectedly.

This paper focuses on combining self-imitation leaning (SIL) \cite{oh2018self} and random network distillation (RND) \cite{burda2018exploration}. SIL is an algorithm that indirectly leads to deep exploration by exploiting only good decisions of the past, whereas RND solves the problem of hard exploration by giving an exploration bonus through deterministic prediction error. The RND bonus is a deterministic prediction error of a neural network predicting features of the observations, and the authors have shown significant performance in some hard exploration Atari games. 
In hard exploration environments, it does not make sense for SIL to exploit a good decision of the past. In other words, SIL requires an intrinsic reward. Meanwhile, in RND, catastrophic forgetting could occur during learning because the predictor network learns about the state that the agent visited recently. Consequently, the prediction error increases, and the exploration bonus increases for previously visited states. We will describe this phenomenon in detail in section 4.3.

This paper introduces \textit{amplifying the imitation effect} (AIE) by combining SIL and RND to drive deep exploration. In addition, we introduce techniques that can enhance the strength of the proposed network.  Adding an intrinsic penalty reward to the state that the agent continuously visits leads to deviation from the current converged policy. Moreover, to avoid catastrophic forgetting, we use a pool of stored samples to update the predictor network during imitation learning such that we can uniformly learn the visited states by the predictor network. We have experimentally demonstrated that these techniques lead to deep exploration.

We verify our algorithm using unmanned combat aerial vehicle (UCAV) mission execution. Some studies have applied RL to UCAV maneuvers. \cite{liu2017deep,zhang2018research,minglang2018maneuvering}. However, those studies simply defined the state and action and experimented in a dense reward environment. We constructed the experimental environment by simulating the flight maneuvers of the UCAV in a three-dimensional (3D) space. The objective of the RL agent is to learn the maneuvers by which the UCAV reaches a target point while avoiding missiles from the enemy air defense network.
The main contributions of this paper are as follows:

\begin{itemize}
\item We show that SIL and RND are complementary and that combining these two algorithms is very efficient for exploration.
\item We  present several techniques to amplify the imitation effect.
\item The performance of the RL applied to the UCAV control problem is excellent. The learning method outputs reasonable UCAV maneuvers in the sparse reward environment.
\end{itemize}

\section{Problem Definition}

\begin{figure}[h]
\vskip 0.2in
\begin{center}
\centerline{\includegraphics[width=1\columnwidth]{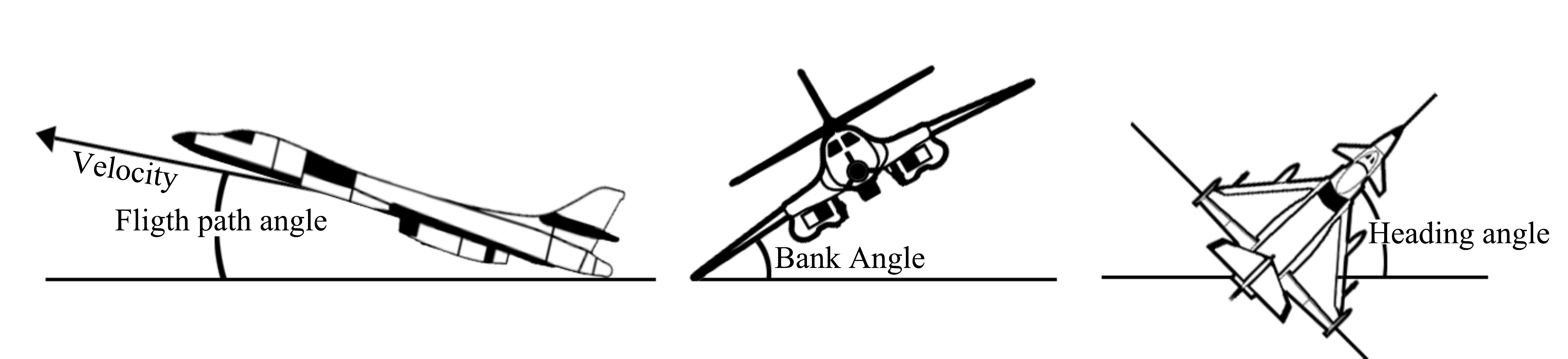}}
\caption{ Bank angle, flight path angle and heading angle of the UCAV. }
\label{ucav}
\end{center}
\vskip -0.3in
\end{figure}

We overlapped the air defense network as in an actual battlefield environment, and we aimed to learn that the UCAV reaches the target by avoiding missiles from the starting point in a limited time period. For the UCAV dynamics, we applied the following equations of motion of a 3-degrees-of-freedom point mass model \cite{kim2007three}:
\begin{eqnarray}
\dot{x} & = & V\cos\gamma\cos\psi \nonumber\\
\dot{y}& = & V\cos\gamma\sin\psi   \nonumber\\
\dot{z} & = & V\sin\gamma    \nonumber\\
\dot{V}& =& {{T-D}\over {m}} - g\sin\gamma  \nonumber\\
\dot{\psi}& =&  {{gn\sin\phi}\over {V\cos\gamma}}   \nonumber\\
\dot{\gamma}& =&  {{g}\over {V(n\cos\phi-\cos\gamma)}}  \label{eq:3dof}
\end{eqnarray}
where ($x$, $y$, $z$) is the position of the UCAV, $V$ is the velocity, $\psi$ is the heading angle, and $\gamma$ is the flight path angle. $T$, $n$ and $\phi$ are the control inputs of the UCAV. $T$, $n$ and $\phi$ denote the engine thrust, load factor and bank angle, respectively. We use these control inputs as the action of our RL framework. Figure \ref{ucav} shows the UCAV's bank angle, flight path angle, and heading angle. The engine thrust affects the velocity of the UCAV. The bank angle and load factor affect the heading angle and flight path angle.

For the missile, we applied proportional navigation induction to chase the UCAV \cite{moran2005three}. We assume that if the distance between the UCAV and the missile is less than 0.5 km, then the UCAV is unable to avoid the missile.

\subsection{State}
In general, in an environment such as Atari games, the image of the game is preprocessed and used as the state, and a convolutional neural network is employed as the structure of the network. In this study, however, the UCAV's coordinate information and the UCAV's radar information to detect missiles are vectorized for the state of the UCAV control problem. A multilayer perceptron is more appropriate for the problem than a convolutional neural network, which is generally adopted for representing an image as the state of an arcade game.

\subsubsection{Coordinate Representation}
In a coordinate system, the coordinate points do not have a linear relationship. For example, the two-dimensional (2D) coordinate (10, 10) is not ten times more valuable than the coordinate (1, 1). Therefore, placing coordinates into a state with real numbers is not reasonable and causes  learning instability. One way to represent the coordinates in the learning environment is to use a one-hot encoding vector. However, the one-hot encoding increases the dimension of the vector as the range of coordinates increases and is only possible for integer coordinates. In this study, we introduce a method to efficiently represent the coordinate system.  \\
The proposed method converts the coordinates into a one-hot encoding vector for each axis and then concatenates the vectors of the axes. The one-hot encoding method requires 40,000 rows (200x200) rows to represent (1, 1) when $x$ and $y$ range from 1 to 200, but using this method, $c_{(1, 1)}=[(1,0,\cdots,0)(1,0,\cdots,0)]'$ is possible with 400 rows (200+200). We additionally extended this method to the real coordinate system. The real coordinates are represented by introducing weight within the vector. For example, 1.3 is close to 70\% in 1 and close to 30\% in 2; in other words, the number 1.3 is a number with a weight of 70\% in 1 and 30\% in 2. Thus, 1.3 can be represented as $c_{(1.3)}=(0.7,0.3,\cdots,0)'$ (200 rows). Moreover, the resulting vector can be reduced to a small dimension. We have reduced this coordinate to 1/10. Consequently, the number 1.3 can be represented as $c_{(1.3)}=(0.13,0,\cdots,0)'$ (20 rows). This method efficiently represents real coordinates within a limited dimension. We call this method efficient coordinate vector (ECV).

\subsubsection{Angle Representation}
\begin{figure}[h]
\vskip 0.2in
\begin{center}
\centerline{\includegraphics[width=1\columnwidth]{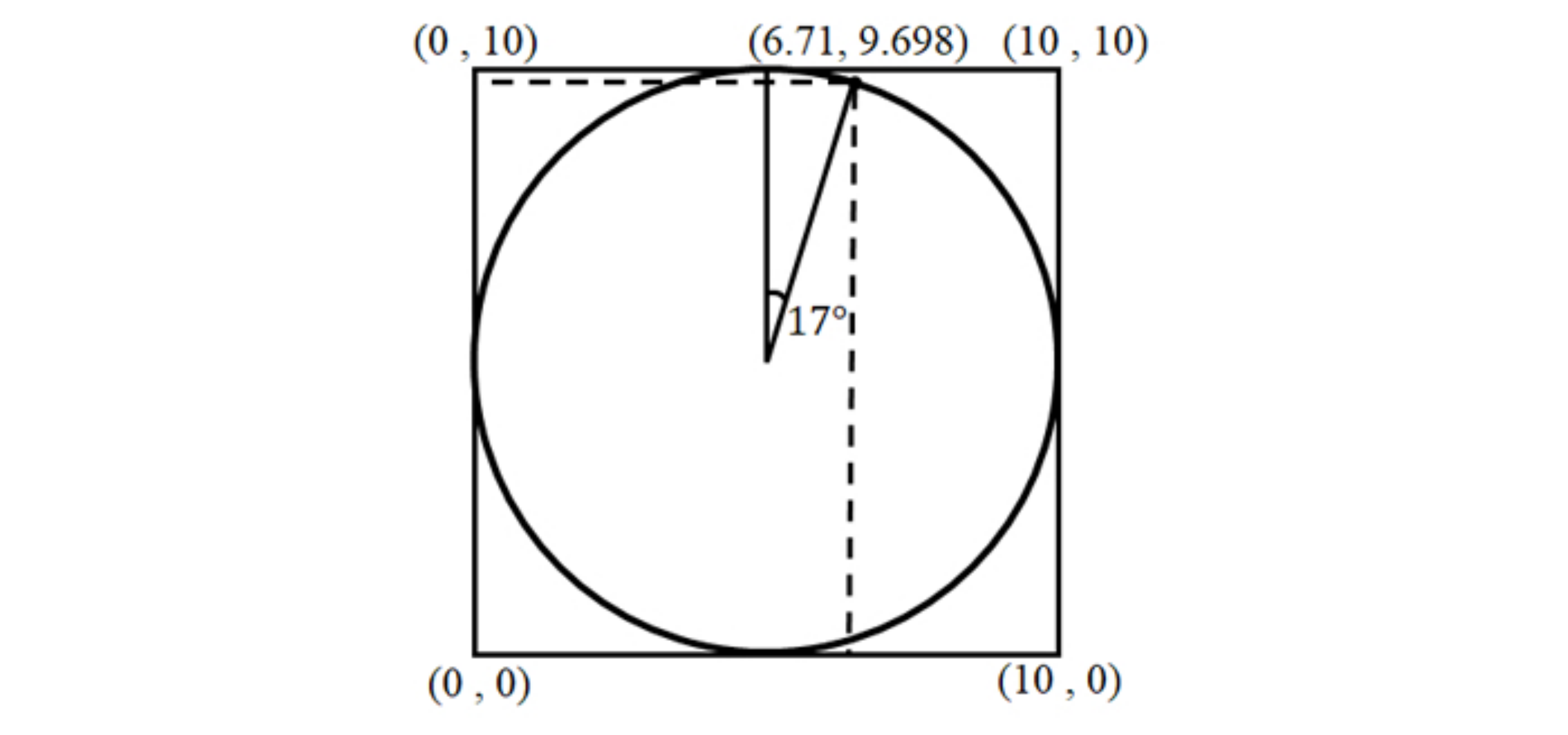}}
\caption{ Example of angle representation. }
\label{angle}
\end{center}
\vskip -0.3in
\end{figure}

Representing the angle as a state is also difficult in RL because the angle has a characteristic of circulating around $360^{\circ}$. For example, suppose that we change the angle from $10^{\circ}$ to $350^{\circ}$. Even if we use a real value or the ECV method, the agent will perceive the result of a $340^{\circ}$ change. However, the difference ($340^{\circ}$) is $20^{\circ}$ at the same time. That is, this angle representation confuses the RL agent. We solve this problem with the polar coordinate system and ECV. $r$ and $\theta$ can be transformed into Cartesian coordinates $x$ and $y$ using a trigonometric function. Using the polar coordinates, we can convert $r$ and $\theta$ into Cartesian coordinates $x$ and $y$. Additionally, we can represent these coordinates as a state through ECV. In other words, the angle is converted into the circle upper position using the polar coordinate system, and then it is represented as a state through the ECV. For example, as shown in figure \ref{angle}, the point on the circle corresponding to $17^{\circ}$ can be represented as $c_{(17^{\circ})}=(0 ,\cdots, 0.71,0.29, 0, \cdots, 0, 0.302,0.698)'$ (20 rows) through ECV.

\subsubsection{Final State}
We finally used the following information as the state of the UCAV control problem.

\begin{itemize}
\item[--] Flight path consisting of five recent steps of the UCAV
\item[--] Path angle, heading angle and bank angle for two recent steps of the UCAV
\item[--] Velocity and load factor of the UCAV
\item[--] Distance between the UCAV and the missile
\item[--] Horizontal and vertical angles between the UCAV and the missile

\end{itemize}

\subsection{Action}

The action is an input combination of engine thrust, bank angle and load factor using Equation 1. Each input has three choices: increase, hold, and decrease. In addition, we have added an action that initializes all inputs to have default values (bank angle: $0^{\circ}$, load factor: $1G$, and engine thrust: $50kN$). This action allows the UCAV to cruise. The total number of actions is 28.

\subsection{Reward}
The default reward is zero, except for the following specific situations:
\begin{itemize}
\item[--] A result of the missile skirmishes 
\item[--] Whether the UCAV has arrived at its target point
\item[--] Cruise condition
\end{itemize}
The cruise condition is rewarded because the UCAV cannot maintain the maximum speed for cruising. We impose a penalty of -0.01 if the speed reaches the maximum speed.


\section{Related Work}
\textbf{Experience replay} Experience replay \cite{lin1992self} is a technique for exploiting past experiences, and Deep Q-Network (DQN) has exhibited human-level performance in Atari games using this technique\cite{mnih2013playing,mnih2015human}. Prioritized experience replay \cite{schaul2015prioritized} is a method for sampling prior experience based on temporal difference.
ACER \cite{wang2016sample}  and Reactor \cite{gruslys2017reactor} utilize a replay memory in the actor-critic algorithm \cite{sutton2000policy,konda2000actor}. However, this method might not be efficient if the past policy is too different from the current policy \cite{oh2018self}. SIL is immune to this disadvantage because it exploits only past experiences that had higher returns than the current value.

\textbf{Exploration} Exploration has been the main challenging issue for RL, and many studies have proposed methods to enhance exploration. Count-based exploration bonus \cite{strehl2008analysis} is an intuitive and effective exploration method in which an agent receives a bonus if the agent visits  a novel state, and the bonus decreases if the agent visits a frequently visited state. There are some studies that estimate the density of a state to provide a bonus in a large state space \cite{bellemare2016unifying,ostrovski2017count,fox2018dora,machado2018count}.
Recent studies have introduced a prediction error (curiosity), which is the difference between the next state predicted and the actual next state for the exploration \cite{silvia2012curiosity,stadie2015incentivizing,pathak2017curiosity,burda2018large,haber2018learning}. The studies designed the prediction error as an exploration bonus ($i_t$) to give the agent more reward when performing unexpected behaviors. 

However, the prediction error has a stochastic characteristic because the target function is stochastic. In addition, the architecture of the predictor network is too limited to generalize the state of the environment. To solve these problems, RND \cite{burda2018exploration} proposed that the target network be deterministic by fixing the network with randomized weights and proposed that the predictor network has the same architecture as the target network. Other methods for efficient exploration include adding parameter noise within the network \cite{strehl2008analysis,plappert2017parameter}, maximizing entropy policies \cite{haarnoja2017reinforcement,ziebart2010modeling}, adversarial self-play \cite{sukhbaatar2017intrinsic} and learning diverse policies \cite{eysenbach2018diversity, gangwani2018learning}.

\textbf{Self-Imitation Learning} SIL can indirectly lead to deep exploration by imitating the good decisions of the past \cite{oh2018self}. To exploit past decisions, the authors used replay buffers $\mathcal{D}$ = \{($s_t,a_t,R_t$)\}, where $s_t$ and $a_t$ are a state and an action at $t$-step, and $R_t= \Sigma^\infty_{k=t}\gamma^{k-t}r_k$ is the discounted sum of reward at $t$-step with a discount factor $\gamma$.  The authors proposed the following off-policy actor-critic loss:

\begin{eqnarray}
\mathcal{L}^{sil}&=&\mathbb{E}_{s,a,R\in \mathcal{D}}[\mathcal{L} _{policy}^{sil} + \beta^{sil}\mathcal{L} _{value}^{sil}] \label{eq:sil1} \\
\mathcal{L}_{policy}^{sil}&=&-log\pi_{\theta}(a|s)(R-V_\theta(S))_+ \label{eq:sil2} \\
\mathcal{L}_{value}^{sil}& =&{{1}\over{2}} \parallel(R-V_\theta(S))_+\parallel ^2 \label{eq:sil3}
\end{eqnarray}

where $(\cdot)_+=max(\cdot,0)$ and $\pi_\theta$ and $V_{\theta}(s)$ are the policy (i.e., actor) and the value function parameterized by $\theta$. $B^{sil} \in \mathbb{R}^+$ is a hyperparameter for the value loss. 
Intuitively, for the same state, if the past return value is greater than the current value ($R>V_\theta$), then it can be observed that the behavior in the past is a good decision. Therefore,  imitating the behavior  is desirable. However, if the past return is less than the current value ($R<V_\theta$), then imitating the behavior is not desirable. The authors focused on combining SIL with advantage actor-critic (A2C) \cite{mnih2016asynchronous} and showed significant performance in experiments with hard exploration Atari games.

\textbf{Random Network Distillation} The authors proposed a fixed target network ($f$) with randomized weights and a predictor network ($\widehat{f}$), which is trained using the output of the target network. The predictor neural network is trained by  gradient descent to minimize the expected mean squared error $\parallel \widehat{f}(x;\theta)-f(x)\parallel^2$. They used the exploration bonus ($i_t$) as $\parallel \widehat{f}(x;\theta)-f(x)\parallel^2$. 
Intuitively, the prediction error will increase for a novel state, and the prediction error will decrease for a state that has been frequently visited. However, if the agent converges to local policy, prediction error may ($i_t$) no longer occurs. Furthemore, using RND can cause catastrophic forgetting. The predictor network learns about the state that the agent constantly visits such that the network forgets about the previously visited state. Consequently, the prediction error increases for the past state, and the agent may go to a past policy.

\section{AIE}

\begin{algorithm}[tb]
   \caption{Amplifying the Imitation Effect (AIE) }
   \label{alg:example}
\begin{algorithmic}
   \STATE Initialize A2C network parameter $\theta_{a2c}$
   \STATE Initialize predictor/target network  parameter $\theta_{p}$, $\theta_{t}$
     \STATE Initialize replay buffer $\mathcal{D} \leftarrow  \emptyset$
   \STATE Initialize episode buffer $\mathcal{E} \leftarrow  \emptyset$
   \STATE Initialize feature buffer $\mathcal{F} \leftarrow  \emptyset$
   \FOR{episode = 1, M}
   \FOR{each step}
   \STATE Execute an action $s_t,a_t,r_t,s_{t+1} \approx \pi_{\theta}(a_t|s_t)$
   \STATE Extract feature of $s_{t+1}$ to $\phi_{s_{t+1}}$
   \STATE Calculate intrinsic reward $i_t$ 
   \IF{ $i_t$ $<$ penalty condition threshold}
   \STATE $i_t  \leftarrow \lambda log(i_t)$
   \ENDIF
   \STATE $r_t=r_t+i_t$
   \STATE Store transition  $\mathcal{E}\leftarrow \mathcal{E} \cup \{(s_t,a_t,r_t)\}$
   \STATE $\mathcal{F} \leftarrow \mathcal{F}\cup \{(\phi_{s_{t+1}},f_{\theta_{t}}(\phi_{s_{t+1}}))\}$
   \ENDFOR
   \IF{ $s_{t+1}$ is terminal}
   \STATE Compute returns $R_t= \Sigma^\infty_{k}\gamma^{k-t}{r}_k$ for all $t$ in $\mathcal{E}$
   \STATE  $\mathcal{D}\leftarrow \mathcal{D}\cup \{(s_t,a_t,r_t)\}$  
   \STATE Clear episode buffer $\mathcal{E} \leftarrow  \emptyset$
\ENDIF
\STATE \bfseries{\emph{ \# Optimize actor-critic network}}
\STATE $\theta_{a2c} \leftarrow  \theta_{a2c} - \eta \nabla_{\theta_{a2c}} \mathcal{L}^{a2c}$ 
\STATE \bfseries{\emph{ \# Perform self-imitation learning}}
  \FOR{k= 1, M}
  \STATE sample a minibatch $\{(s,a,R)\}$ from $\mathcal{D}$
  \STATE  $\theta_{a2c} \leftarrow  \theta_{a2c} - \eta \nabla_{\theta_{a2c}} \mathcal{L}^{sil}$ 
  \STATE sample a minibatch $\{(\phi_{s_{t+1}},f_{\theta_{t}}(\phi_{s_{t+1}}))\}$ from $\mathcal{F}$
  \STATE  $\theta_{p} \leftarrow  \theta_{p} - \eta \nabla_{\theta_{p}} \mathcal{L}^{p}$ 
 \ENDFOR
\ENDFOR

\end{algorithmic}
\label{algo1}
\end{algorithm}

\subsection{Combining SIL and RND}
In this section, we explain why combining RND and SIL can amplify the imitation effect and lead to deep exploration. The SIL updates only when the past $R$ is greater than the current $V_\theta$ and imitates past decisions. Intuitively, if we combine SIL and RND, we find that the ($R-V_\theta$) value is larger than the SIL because of the exploration bonus. In the process of optimizing the actor-critic network to maximize $R_t= \Sigma^\infty_{k=t}\gamma^{k-t}{(i_t+e_t)}_k$, where $i_t$ is intrinsic reward and $e_t$ is extrinsic reward, the increase in $i_t$ by the predictor network causes $R$ to increase. That is, the learning progresses by weighting the good decisions of the past. This type of learning thoroughly reviews the learning history.
 If the policy starts to converge as the learning progresses, the $i_t$ will be lower for the state that was frequently visited. One might think that learning can be slower as $(R_{t}-V_\theta) > (R_{t+k}-V_\theta)$, where $k > 0$ for the same state and $i_t$ decreases. However, the SIL exploits past good decisions and leads to deep exploration. By adding an exploration bonus, the agent can further explore novel states. Consequently, the exploration bonus is likely to continue to occur. In addition, using the prioritized experience replay \cite{schaul2015prioritized}, the sampling probability is determined by the ($R-V_\theta$); thus, there is a high probability that the SIL will exploit the previous transition even if $i_t$ decreases. In other words, the two algorithms are complementary to each other, and the SIL is immune to the phenomenon in which the prediction error ($i_t$) no longer occurs.

\subsection{Intrinsic Penalty Reward}		
Adding an exploration bonus to a novel state that the agent visits is clearly an effective exploration method. However, when the policy and predictor networks converge, there is no longer an exploration bonus for the novel state. In other words, the exploration bonus method provides a reward when the agent itself performs an unexpected action, not when the agent is induced to take the unexpected action. Therefore, an exploration method that entices the agent to take unexpected behavior is necessary. We propose a method to provide an intrinsic penalty reward for an action when it frequently visits the same state rather than rewarding it when the agent makes an unexpected action. The intrinsic penalty reward allows the agent to escape from the converged local policy and helps to experience diverse policies. 
Specifically, we provide a penalty by transforming the current intrinsic reward into $\lambda log(i_t) $, where $\lambda$ is a penalty weight parameter, if the current intrinsic reward is less than the quantile $\alpha$ of the past $N$ intrinsic rewards. This reward mechanism prevents the agent from staying in the same policy. In addition, adding a penalty to the intrinsic reward indirectly amplifies the imitation effect. Since the $(R_{t}-V_\theta)$ becomes smaller due to the penalty, the probability of sampling in replay memory is relatively smaller than that of non-penalty transition. 
SIL updates are more likely to exploit non-penalty transitions. Even if $(R_{t}-V_\theta) < 0$ due to a penalty, it does not affect SIL because it is not updated because of the objective of SIL in equation \ref{eq:sil3}. In other words, the intrinsic penalty reward allows the policy network to deviate from the constantly visited state of the agent and indirectly amplifies the imitation effect for the SIL.

\subsection{Catastrophic Forgetting in RND}
The predictor network in RND mainly learns about the state that the agent recently visited, which is similar to the catastrophic forgetting of continual task learning that forgets learned knowledge of previous tasks. If the prediction error increases for a state that the agent has visited before, the agent may recognize the previous state as a novel state. Consequently, an agent cannot effectively explore. The method to mitigate this phenomenon is simple but effective. We store the output of the target network and state feature as the memory of the predictor network, just like using a replay memory to reduce the correlation between samples\cite{mnih2013playing}, and train the predictor network in a batch mode. Using the predictor memory reduces the prediction error of states that the agent previously visited, which is why the agent is more likely to explore novel states. Even if the agent returns to a past policy, the prediction error of the state visited by the policy is low, intrinsic penalty is given to the state, and the probability of escaping from the state is high.

\begin{figure}[t]
\vskip 0.1in
\begin{center}
\centerline{\includegraphics[width=1\columnwidth]{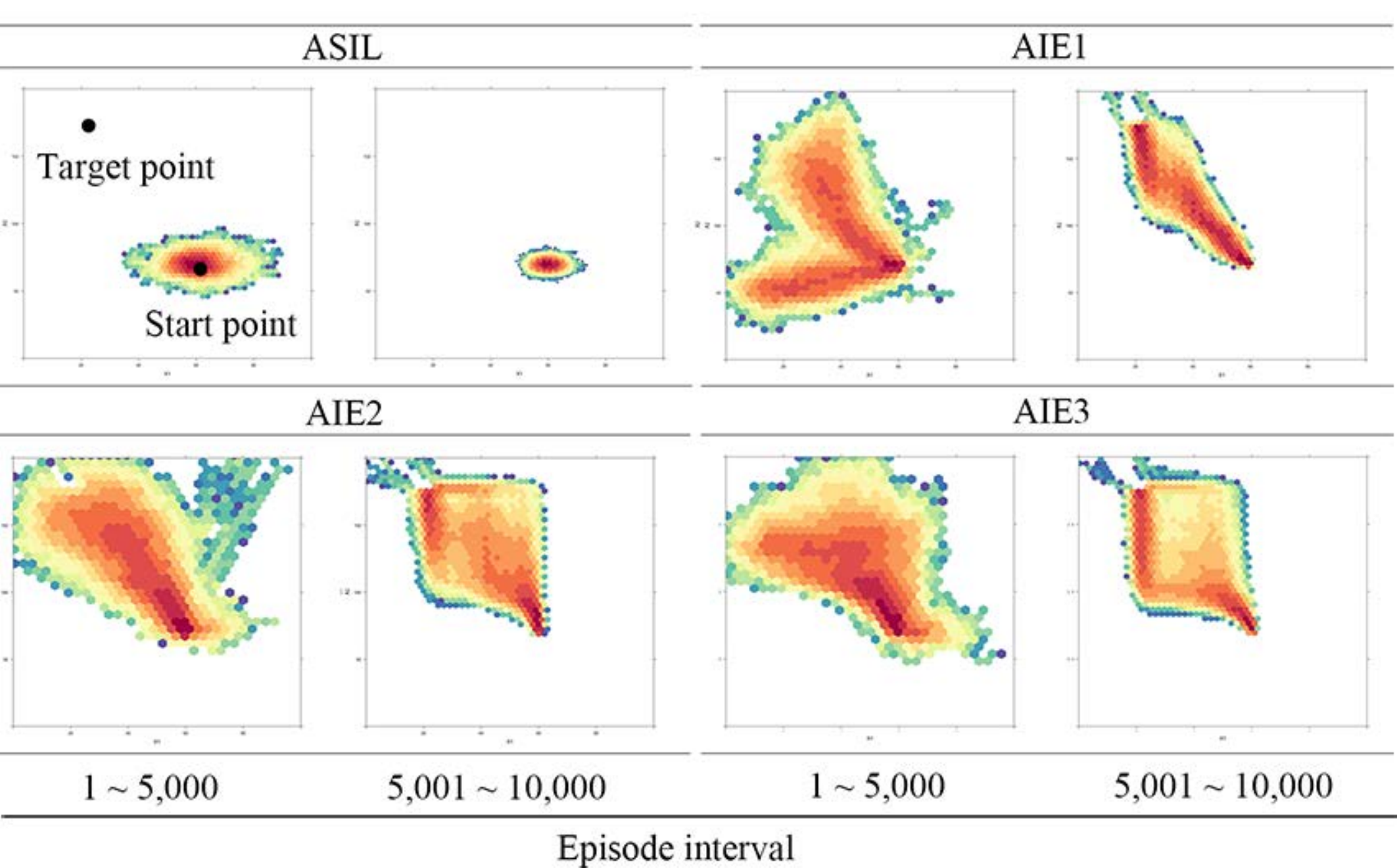}}
\caption{Path visualization for each algorithm in 2D grid environment.  The color changes from blue to red for where the agent visits more frequently.}
\label{exp1}
\end{center}
\vskip -0.4in
\end{figure}

\section{Experiment}

\subsection{Conversion of State to Coordinate Feature}
An exploration bonus is given for state feature $x$ through $\parallel \widehat{f}(x;\theta)-f(x)\parallel^2$, where ($f$) is a fixed target network and ($\widehat{f}$) is a predictor network. However, the state of our experimental environment contains various information, such as the path and direction information of the UCAV and the relationship information between the UCAV and missile. The high-dimensional state space makes the convergence speed of the policy network slow. Thus, we limited the state for the exploration bonus to the current coordinates of the UCAV (33 rows). Consequently, the convergence rate of the policy network increased, and the meaning of the role of the exploration bonus changes clearly from ‘inducing the agent to move to a novel feature state’ to ‘inducing agent to move to novel coordinates’.

\subsection{Test Algorithms}
ASIL denotes the combination of A2C and SIL. We used this model as a baseline method for a performance comparison. In this study, we propose three RL algorithms. Amplifying the imitation effect (AIE1) is the first proposed algorithm, which combines ASIL and RND. The second is the addition of intrinsic penalty rewards to ASIL + RND (AIE2), and the third is the AIE2 with the addition of replay memory for the predictor network (AIE3) described in Algorithm \ref{algo1}.

\begin{figure*}[t]
 \center
  \includegraphics[width=\textwidth]{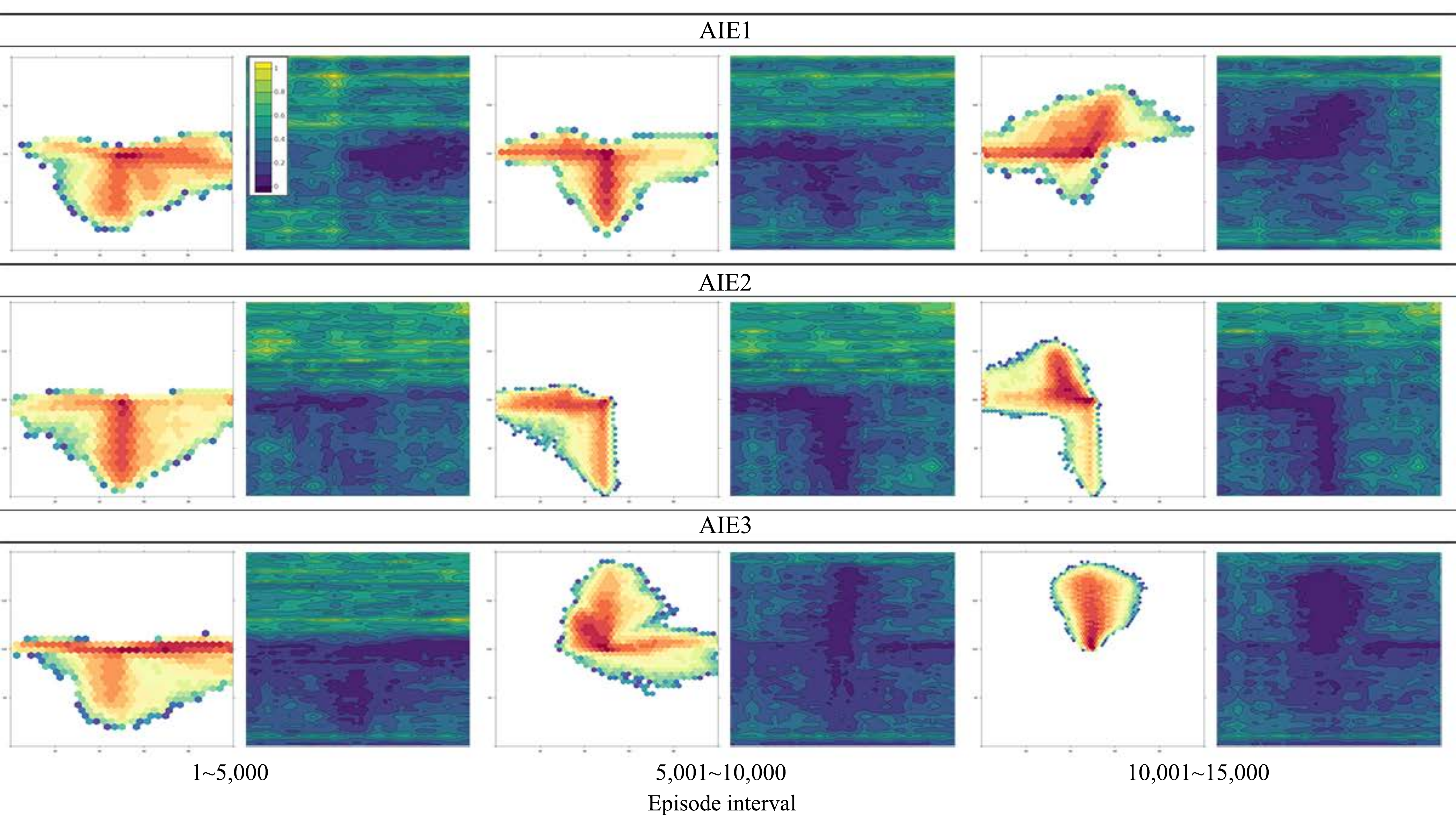}
  \caption{
Visualization of the path of the agent and loss of all coordinate states for each algorithm in the no reward 2D grid environment.  The color changes from blue to red in the agent's path figure to indicate where the agent visits more frequently.  The color changes from blue to yellow in the loss figure to indicate where the loss is larger.}
  \label{exp2}
\vskip -0.2in
\end{figure*}
\subsection{Hard Exploration in 2D Environment}

\subsubsection{Sparse Reward Setting}
We conducted a simple experiment to see how effective the proposed algorithms are for exploration. We constructed a 2D grid world in which the agent learns a sequence of movements that begin from a starting point and reach a goal point using a simple movement step (up, down, left, and right). The reward was set to zero except when reaching the target point (reward of 30) or leaving the environment (reward of -30). RL was performed a total of 10,000 episodes for each algorithm. Figure 2 is the visualization of the movement paths of the agent. Since the reward is too sparse, the ASIL failed to reach the target point. In contrast, all of the proposed algorithms successfully reached the target point because of the exploration bonus. For  AIE1, the result showed that the agent quickly reached the target point. However, we find that AIE 2 and AIE3 that considered the intrinsic penalty reward performed a deeper exploration than AIE1 – the two algorithms arrived at the target point via more diverse paths compared to AIE1.

\begin{table}[t]
\caption{An exploration area score of each algorithm in a two-dimensional no-reward grid environment. We averaged the area explored by the agent after 30 repeated experiments.}
\label{table}

\begin{center}
\begin{small}
\begin{sc}
\begin{tabular}{lcccr}
\toprule
Algorithm & Exploration area \\
\midrule
ASIL    & 11.2$\pm1.25$ \\
AIE1    & 40.5$\pm2.06$  \\
AIE2 & 43.2$\pm2.36$ \\
AIE3  & 46.7$\pm2.19$  \\
\bottomrule
\end{tabular}
\end{sc}
\end{small}
\end{center}
\end{table}

\subsubsection{No-Reward Setting}
We experimented with the same environment in which there is no target point. The agent performs only exploration in each episode. We argue that the catastrophic forgetting is ineffective for exploration when using an exploration bonus because the agent has less chance of searching a novel state if the prediction error remains high for previously searched states. Furthermore, we argue that using replay memory for predictor network (AIE3) is more efficient for exploration because the memory mitigates the catastrophic forgetting.

Figure \ref{exp2} is the visualization of the movement paths of the agent for 5,000 episodes (left figure) and the losses of the predictor network at all coordinates (right figure). We observed that the loss of the area explored by the agent is lower than in other areas. As the episode increases, the agent explores  a novel space with a high prediction error. At this point, we can observe that the loss of  area that the agent explored at an episode increased compared to the loss of area at the preceding episode. However, AIE3 showed that the loss of the previously explored space remained relatively low compared to the other two algorithms.

In the sparse reward environment,  ASIL explored a small area, circulating throughout the area although the episode increased, but the proposed three algorithms explored many areas. Table \ref{table} shows the score of how each algorithm explored uniformly over four quadrants of the 2D grid space during 30,000 episodes. The formula for the score was 
 \begin{eqnarray}
score& = & mean(EQ_{q}) \times \sigma_{EQ} \times 100 \label{eq:score}
\end{eqnarray}
where $EQ_{q}$ is the explored portion in the total area of each quartile. We confirmed that the proposed algorithms (particularly AIE3) were very effective for exploration.

\begin{figure}[h]
\vskip 0.2in
\begin{center}
\centerline{\includegraphics[width=1\columnwidth]{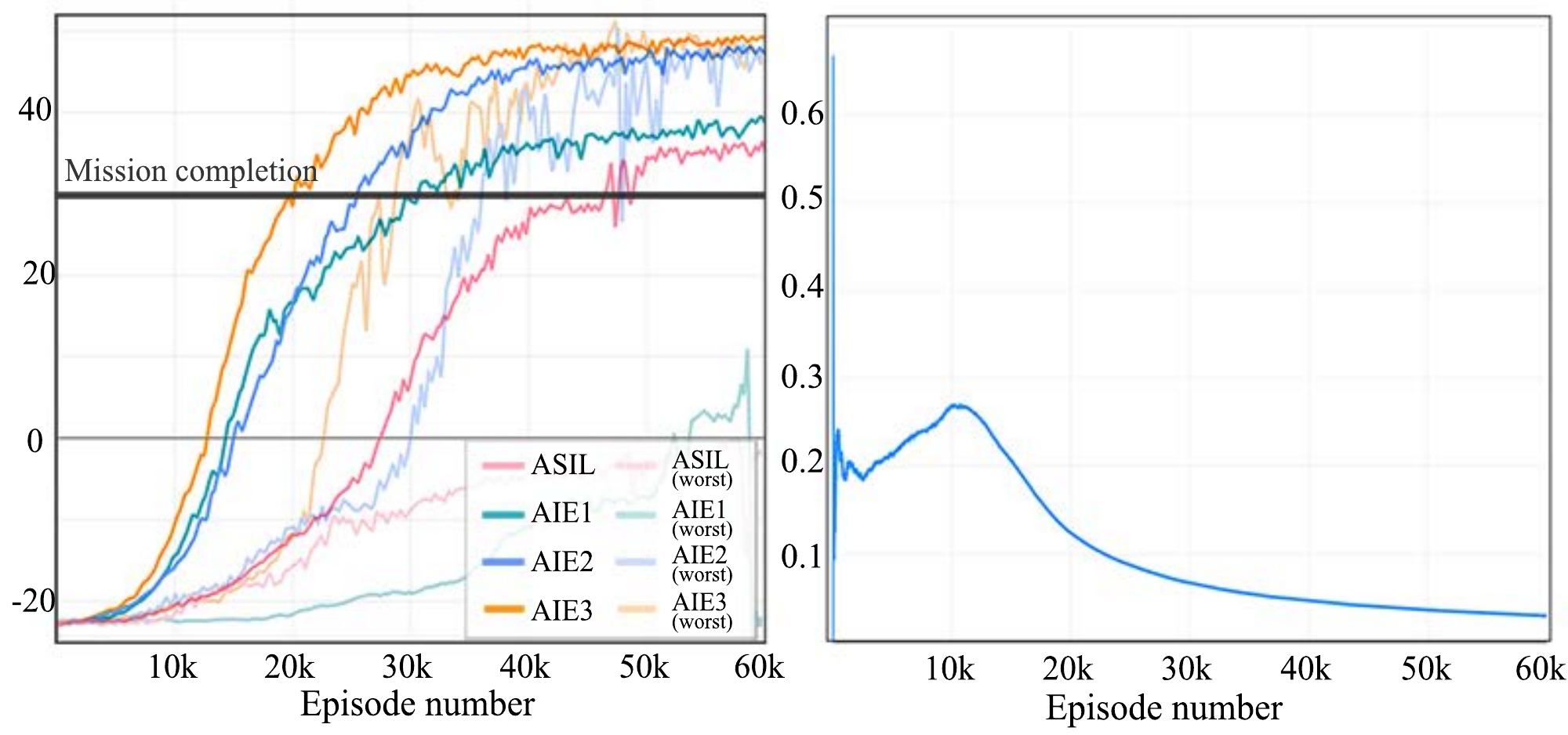}}
\caption{(Left) Learning curves of the UCAV mission execution environment. The x and y axes represent the episode number and the average reward, respectively. The plot is the average of the reward of the results of the 10 experiments for each algorithm. The light color represents the worst performance result of each algorithm. (Right) Cumulative probability graph of being shot down by a missile. }
\label{ucavresult}
\end{center}
\vskip -0.3in
\end{figure}

\begin{figure}[t]
\vskip 0.2in
\begin{center}
\centerline{\includegraphics[width=1\columnwidth]{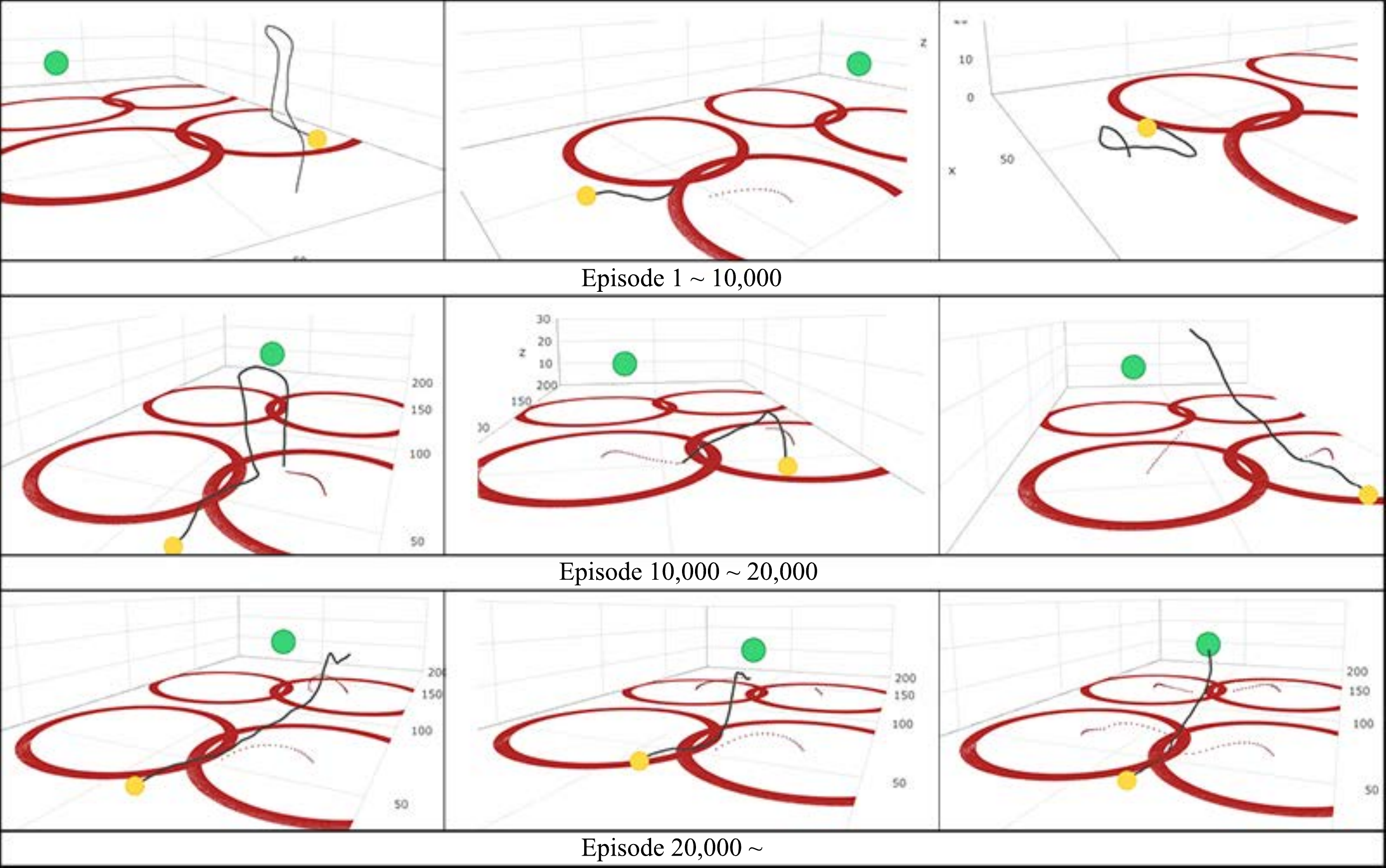}}
\caption{3D view of the UCAV's learning process. The red circle represents the air defense network, the black solid line represents the movement path of the UCAV, and the red dotted line represents the missile's movement path.}
\label{process}
\end{center}
\vskip -0.2in
\end{figure}

\subsection{Experiment for UCAV Mission Execution}
We performed an experiment to investigate  UCAV control in a sparse reward environment and compared the performances of the algorithms. In addition, we analyzed how the UCAV manages to avoid missiles. First, since our experimental environment has a sparse reward structure, DQN, prioritized experience replay DQN, A2C and ACER failed to converge to the desired policy that generates the shortest path from the origin to the target point while avoiding an enemy’s missiles. Figure \ref{ucavresult} (left) shows the performances of ASIL and the proposed three algorithms for an experiment consisting of 60,000 episodes. 
The light colors and normal colors represent the worst and average performance of the compared algorithms, respectively. 
The result is that AIE2 and AIE3 succeeded in converging to the desired policy, while ASIL and AIE1 fell into a local minimum once in two trials and once in three trials, respectively. In particular, AIE3 outperformed the other algorithms, as shown in Figure 4. Similar to the previous exploration experiment, we confirmed that the performance of the three proposed algorithms was better than that of ASIL (baseline model) in the UCAV control environment.

Figure \ref{process} presents snapshots of learning (animation is here\footnote{https://youtu.be/7R5lZAsCs2c}). At early episodes of the learning, the UCAV took random actions and occasionally left the battlefield. However, as the episodes increased, it tended to move forward gradually but was shot down by a missile. This result can be confirmed by the cumulative shot probability plot (Figure \ref{ucavresult} (right)). As the episodes continues, the UCAV learned how to avoid missiles and began to move to new coordinates (attempted to increase intrinsic reward). The UCAV attempted to reach the target point through various paths.

\begin{figure}[h]
\vskip 0.2in
\begin{center}
\centerline{\includegraphics[width=1\columnwidth]{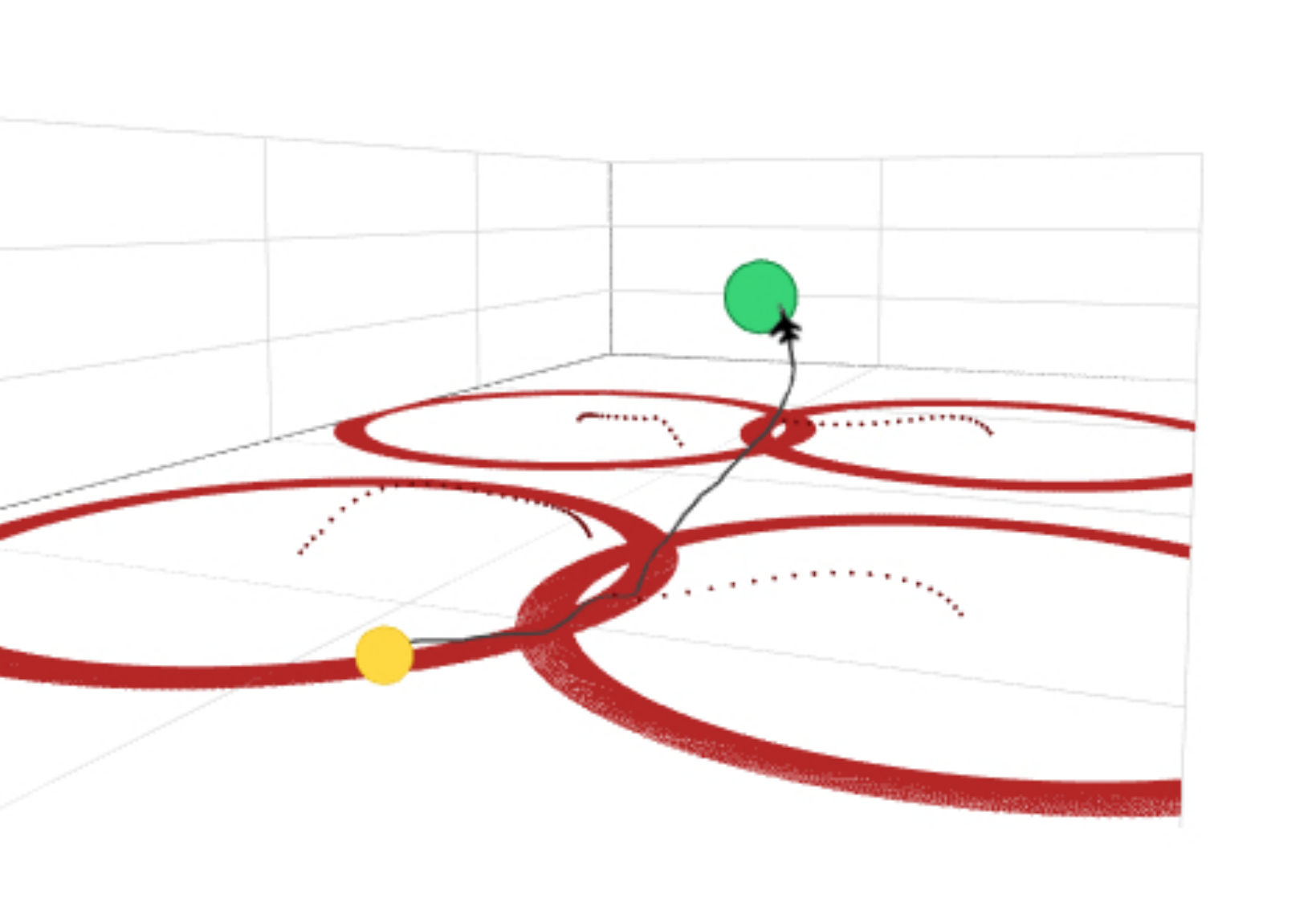}}
\caption{UCAV's path of after learning in 3D view. You can see the UCAV through the overlapping of air defense network, avoiding the missile and reaching the target point.}
\label{ucavroute}
\end{center}
\vskip -0.3in
\end{figure}

Figure  \ref{ucavroute} is a 3D representation of the path through which the UCAV reached the target while avoiding the missile. When the UCAV entered the center of the air defense network, the probability of being shot down by a missile increased. Therefore, the UCAV learned the safe path that passed through the overlapped areas of air defense networks with a low altitude.


\section{Conclusion}
In this paper, we proposed  AIE by combining SIL and RND. In addition, we proposed AIE2 and AIE3, which can lead to efficient deep exploration. AIE2 gives an intrinsic penalty reward to states where the agent frequently visits, which prevents the agent from falling into a local optimal policy. AIE3 adopts replay memory to mitigate the catastrophic forgetting of the predictor network. These two algorithms amplify the imitation effect, leading to deep exploration, thereby enabling the policy network to quickly converge into the desired policy. We experimentally demonstrated that the AIEs in the 2D grid environment successfully explored wide areas of the grid space. In addition, for the UCAV control problem, we observed that the proposed algorithms quickly converged into the desired policy. In  future work, it is necessary to discuss the configuration of the replay memory because replay memory for the predictor network has  limited storage; thus, it is inefficient to insert a feature for every learning step.

\section*{Acknowledgments}
This research was supported by Agency for Defense Development (UD170043JD).


\bibliography{reference}
\bibliographystyle{icml2019}



\end{CJK}
\end{document}